\documentclass{article} 
\usepackage{nips15submit_e,times}
\usepackage{hyperref}
\usepackage{url}
\usepackage{amsmath}
\usepackage{amssymb}
\usepackage{amsthm}
\usepackage{algorithm}
\usepackage{algorithmic}
\usepackage{graphicx}

\usepackage{caption}
\usepackage{subcaption}

\title{A Dictionary Learning Approach for Factorial Gaussian Models }

\author{
Y. Cem S\"{u}bakan$^\flat$, Johannes Traa$^\sharp$, Paris Smaragdis$^{\flat,\sharp,\natural}$, Noah Stein$^{\sharp\sharp}$\\
$^\flat$Department of Computer Science, University of Illinois at Urbana-Champaign \\
$^\sharp$Department of Electrical and Computer Engineering, University of Illinois at Urbana-Champaign \\
$^\natural$Adobe Systems, Inc. $^\natural\natural$Analog Devices \\
\texttt{ \{subakan2, traa2, paris\}@illinois.edu, noah.stein@analog.com} \\
}

\DeclareMathOperator*{\argmin}{arg\,min}

\nipsfinalcopy 

\begin{document}

\maketitle

\begin{abstract}
In this paper, we develop a parameter estimation method for factorially parametrized models such as Factorial Gaussian Mixture Model and Factorial Hidden Markov Model. Our contributions are two-fold. First, we show that the emission matrix of the  standard Factorial Model is unidentifiable even if the true assignment matrix is known. Secondly, we address the issue of identifiability by making a one component sharing assumption and derive a parameter learning algorithm for this case. Our approach is based on a dictionary learning problem of the form $X = O R$, where the goal is to learn the dictionary $O$ given the data matrix $X$. We argue that due to the specific structure of the activation matrix $R$ in the shared component factorial mixture model, and an incoherence assumption on the shared component, it is possible to extract the columns of the $O$ matrix without the need for alternating between the estimation of $O$ and $R$. 
\end{abstract}

\newcommand{\tran}{A}
\newcommand{\ems}{O}
\newcommand{\hmmK}{M}
\newcommand{\hmmL}{L}
\newcommand{\nhmm}{K}
\newcommand{\ind}{R}
\newcommand{\obs}{x}
\newcommand{\dictel}{\mu}
\newcommand{\diag}{\text{diag}}

\section{Introduction} 
In a typical Gaussian Mixture Model (GMM), each data item is associated with a single Gaussian mean, which assumes that only a single cause is active for each observation. While a GMM may be appropriate for some applications such as clustering, it is not expressive enough for modeling data which possess dependency on multiple variables. 

In a factorial representation of hidden state variables, each data item is dependent on $\nhmm > 1 $ variables, where each of which is chosen according to a separate hidden variable \cite{saund95, ghahramani95fact, hinton94}. In the case where the state variables are independent, we call this model a \textit{Factorial Mixture Model}. If there exists a first order temporal dependence between the state variables, it becomes the well-known \textit{Factorial Hidden Markov Model} (FHMM) \cite{ghahramani97fhmm, ghahramani01hmmreview}. Factorial HMMs have found use in numerous unsupervised learning applications such as source separation in audio processing \cite{mysore12}, de-noising in speech recognition \cite{virtanen06}, vision \cite{facehmm}, and natural language processing \cite{duh05fhmm}.  

Although factorial models have been used extensively in practice, parameter learning is mainly limited to search heuristics such as variational Expectation Maximization (EM) algorithm and Markov Chain Monte Carlo (MCMC) methods \cite{ghahramani97fhmm, ghahramani01hmmreview},  which iterate between the coordinate-wise updates for each parameter until local convergence. These methods require good initializations and require an indefinite number of iterations. 


In this paper we have two main contributions. We first show that it is impossible to recover the true emission matrix of a factorial model even if we have the true assignment matrix. We then present an algorithm which finds a global solution under incoherence and one-component sharing assumptions for the emission parameters.

\subsection{Notation}
 We use the MATLAB colon notation $A(:,j), A(j,:)$, which in this case respectively picks the $j$'th column/row of a matrix $A$. We use the subscript notation $x_{1:T}$ to denote $\{ x_1, x_2, \dots, x_{T} \}$. A probability simplex in $\mathbb R^N$ is denoted by $\Delta^{N-1} : = \{ (p_1, p_2, \dots, p_N)\in \mathbb R^{N} :p_i \geq 0 \; \forall i, \; \sum_{i=1}^{N} p_i =1 \} $. We denote the space of column stochastic matrices of size $N \times M$ with $\Delta^{N-1 \times M}$. An indicator function is denoted by $\mathbf 1(arg)$: If $arg$ is true then the output is 1, otherwise the output is zero. For a positive integer $N$, let $[N]:=\{ 1,\dots, N\} $. We also use square brackets to concatenate matrices: Let $A\in \mathbb R^{I \times J}$, $B\in \mathbb R^{I \times J}$, then $[A \;  B] \in \mathbb R^{I \times 2J}$. The $N$ dimensional indicator vector is denoted by $e_i \in \mathbb R^N$, where only the $i$'th entry is one and the rest is zero. All-zeroes and all-ones vectors of length $N$ are respectively denoted by $0_N$ and $1_N$. The element wise multiplication of matrices $A$ and $B$ is denoted by $A \cdot B$. For two vectors $a,b\in \mathbb R^N$, we denote the inner product operation with $\langle a,b \rangle = a^\top b$. The identity matrix in $\mathbb R^{N\times N}$  is denoted by $I_N$. 

\subsection{Definitions and Background}
\noindent \textbf{Gaussian Mixture Model (GMM):}
In a GMM, observations $\obs_t \in \mathbb R^\hmmL$ are generated conditioned on latent state indicators $r_t \in e_{1:M}$, such that 
\begin{align}
\obs_t = O r_t+ \epsilon , \; \forall t \in [T], 
\end{align}
where the latent state indicators are i.i.d, $\text{Pr}(r_t = e_i) = \pi_{i}$, $\forall t \in [T]$, $O = \mathbb E[\obs_t | r_t ] \in \mathbb R^{\hmmL \times \hmmK}$ is the emission matrix, and $\epsilon$ is a zero mean Gaussian noise with covariance matrix $\Sigma \in \mathbb R^{\hmmL \times \hmmL}$. Although $\epsilon$ may depend on the cluster indicator $r_t$ in the general case, we show it to be fixed in this equation to make the transition to a factorial model clearer.  

\noindent \textbf{Factorial Gaussian Mixture Model (F-GMM):}
Different from a GMM, in an F-GMM an observation is conditioned on a collection of state variables $\ind_t = [(r_t^1)^\top, (r_t^2)^\top, \dots,(r_t^\nhmm)^\top]^\top$, where $r_t^k \in e_{1:M^{(k)}}$, $\forall k \in [K]$. Without loss of generality, to keep the notation uncluttered we assume that $M^{(k)} = M$, $\forall k \in [K]$. An observation $\obs_t$ is the sum of $K$ vectors chosen by $\ind_t$:
\begin{align}
\obs_t = [O^1, O^2,\dots,O^\nhmm] \ind_t + \epsilon, \;  \forall t \in [T], \label{eq:emsgmm}
\end{align}    
where $O^k \in \mathbb R^{\hmmL \times \hmmK}$, $\epsilon \sim \mathcal N(0,\Sigma)$, and $\text{Pr}(r_t^k = e_i) = \pi_{i}^k, \; \forall t \in [T]$. We denote the \emph{assignment matrix} formed by $R_{1:T}$ with $R \in \mathbb R^{\hmmK \nhmm \times T}$. 

\noindent \textbf{Factorial Hidden Markov Model (F-HMM):}
The only difference between an F-HMM and F-GMM is the dependency structure of the latent state indicators. In an F-HMM, state indicators $r_t^k$ are not independent, but have a Markovian dependency such that $\text Pr(r_{t+1}^k =e_i | r_{t}^k=e_j ) = A^k_{i,j}$, where $A \in \Delta^{\hmmK-1\times \hmmK}$ is the transition matrix of the $k$'th chain. The observation model is exactly the same as F-GMM, and is given by Equation \eqref{eq:emsgmm}.

The proposed learning algorithm in this paper is mainly based around the \textit{dictionary learning} problem of the form $X = O R + \epsilon$, where the dictionary matrix (or the emission matrix) $O = [O^1, O^2,\dots,O^\nhmm] \in \mathbb R^{\hmmL \times \nhmm\hmmK}$ is composed of concatenations of individual dictionaries, and the assignment matrix $R \in \mathbb R^{\hmmK\nhmm \times T}$ consists of $K$ sparse vectors $R_{1:T}$. Given the data matrix $X \in \mathbb R^{\hmmL \times T}$, the learning goal is to estimate the dictionary matrix  $[O^1, O^2,\dots,O^\nhmm]$, upto permutation of the columns within each block $O^{k}$, and upto permutation of the blocks. 

\noindent \textbf{Background:}
The naive approach for dictionary learning is based on \textit{alternating minimization}. The basic idea is to alternate between the estimation of the dictionary and the assignment matrix until convergence. Examples include \cite{Lewicki00, k-svd, engan99}. These approaches are only guaranteed to converge locally.   

There exist only few algorithms in the literature which can estimate the dictionary matrix without an alternating minimization scheme. In \cite{erspud}, an exact recovery algorithm is proposed. The proposed algorithm requires the assignment matrix to be sparse and to have a norm preserving property, and the dictionary matrix to be square. In \cite{anima2012a, anima2012b, hsu13} global algorithms are proposed for learning latent variable models, which correspond to the cases where the columns of $R$ are 1-sparse indicator vectors. Consequently, these algorithms cover the GMM case but not F-GMMs and F-HMMs. More recently, an algorithm based on computing pairwise correlations between observations to find overlapping components is proposed in \cite{moitra14}. The algorithm requires all of the dictionary elements to be incoherent from each other, which may be limiting in our case. Our algorithm is similar in the sense that it uses correlations to extract the components. However, it is based on the specific correlation structure of the factorial models that we work on.  


\section{Identifiability}
\label{sec:identify}

As stated earlier, the learning goal is to estimate the dictionary matrices $O^k = [\dictel^k_1, \dictel^k_2, \dots, \dictel^k_\hmmK ], \; \forall  k\in [K]$, upto permutation of the columns $\dictel^k_{1:\hmmK}$ of each dictionary, and upto permutation of the dictionaries. We assume that the individual emission matrices have full column rank $\textbf{rank}(O^k) =\hmmK$. Unfortunately the emission matrix of a Gaussian factorial model in its original form in \cite{ghahramani95fact, ghahramani97fhmm, ghahramani01hmmreview} is unidentifiable: Even if an oracle gives the true assignment matrix $R$, there are infinitely many plausible dictionary matrices $O$. We will show that the assignment matrix $R$ is rank deficient, which will lead us to the conclusion of unidentifiability. 

\newtheorem{lem}{Lemma}
\newtheorem{cor}{Corollary}
\newtheorem{defn}{Definition}
\newtheorem{thm}{Theorem}

\begin{lem}
\label{lem:rank}
 {Let $R^c \in \mathbb R^{\hmmK\nhmm\times \hmmK^\nhmm}$ denote a matrix whose columns consist of all possible combinations $\ind_t$ can take (e.g. for $\hmmK=2$, $\nhmm =2$ case $R^c = \begin{bmatrix} e_1& e_1 &e_2 & e_2 \\ e_1 & e_2 & e_1 & e_2 \end{bmatrix}$ ). We conclude that $\textbf{rank}(R^c) =  \hmmK\nhmm - (\nhmm -1) $.} 
\end{lem}

\newcommand{\indf}{r}
\noindent \textbf{Proof:} 
We will show this by computing the dimensionality of the left null space of $R^c$. Let, 
$$\indf^k_m(m_1, m_2,\dots,m_k,\dots,m_\nhmm) :=\left\{
     \begin{array}{lr}
       1, &  \text{if} \; m_k = m \\
       0,  &  \text{otherwise}  
     \end{array},
   \right. $$ 
where $k \in [K]$, and $m \in [M]$. This function returns the $(k-1)\hmmK + m$'th row of the column of $R^c$ that corresponds to the combination represented by the tuple $(m_1, m_2,\dots,m_k,\dots,m_\nhmm)$, where $m_k \in [M]$. For a vector $\alpha \in \mathbb R^{\hmmK \nhmm} \in \textbf{null}( (R^c)^\top)$, by definition $\alpha^\top R^c =0_{\hmmK^\nhmm}^\top$. Let us consider the structure of such $\alpha$:
 \begin{align}
 \sum_{k=1}^\nhmm \sum_{m=1}^\hmmK \alpha^k_m r^k_m(m_1, m_2,\dots,m_k,\dots,m_\nhmm) = \sum_{k=1}^\nhmm  \alpha^k_{m_k} = 0  \label{eq:rankpf1}
\end{align}
So, we see that the sum of the elements $\alpha$ that correspond to different $k$'s should sum up to zero. Furthermore for a tuple that only differs in $k$'th element: 
\begin{align}
 \sum_{k=1}^\nhmm \sum_{m=1}^\hmmK \alpha^k_m r^k_m(m_1, m_2,\dots,\widetilde{m}_k,\dots,m_\nhmm) =\sum_{\substack{ k' \neq k} }  \alpha^{k'}_{m_{k'}} + \alpha^k_{\widetilde{m}_k} = 0, \label{eq:rankpf2}
\end{align}
where $m_k \neq \widetilde{m}_k$, and $\forall k \in [K]$. By comparing Equations \eqref{eq:rankpf1} and \eqref{eq:rankpf2}, we see that $\alpha^k_{\widetilde{m}_k} = \alpha^k_{m_k} $. And consequently $\alpha^k_{m} = \alpha^k_{m'}$, $\forall (m,m') \in [M]$, and $\forall k \in [K]$. Together with the constraint $\sum_{k=1}^K \alpha^k_{m_k} =0$ , we conclude that $\textbf{dim}(\textbf{null}((R^c)^\top)) = \nhmm -1$. Therefore, from the rank-nullity theorem, $\textbf{rank}(R^c) = \nhmm \hmmK - \textbf{dim}(\textbf{null}((R^c)^\top)) = \nhmm \hmmK -  (\nhmm -1)$. $\hfill \square$ \\

\begin{cor}
\label{cor:rankR}
The rank of the assignment matrix $R\in\mathbb R^{\hmmK \nhmm \times T}$ is upper bounded: $\textbf{rank}(R) \leq \nhmm \hmmK -  (\nhmm -1)$. 
\end{cor}

\noindent \textbf{Proof:} 
The columns of the assignment are such that $R_t = R^c e_l,\; l \in[\hmmK^\nhmm]$. If $R$ happens to contain all columns of $R^c$, it achieves the rank of $R^c$. In the case where $R$ does not contain all columns of $R^c$, its rank is smaller than $\nhmm\hmmK -  (\nhmm -1)$. Therefore, $\textbf{rank}(R) \leq \nhmm \hmmK -  (\nhmm -1)$. $\hfill \square$ \\


%


\begin{thm}
{Given an assignment matrix $R \in \mathbb R^{\nhmm\hmmK \times T}$, the emission matrix of a Gaussian factorial model is not  identifiable, meaning there exists $O_1\neq O_2 \in \mathbb R^{\hmmL \times \nhmm \hmmK}$ such that $\prod_{t=1}^T \mathcal N (x_t|O_1\ind_t, \Sigma) = \prod_{t=1}^T \mathcal N(x_t |O_2 \ind_t, \Sigma)$. }\\

\end{thm}

\noindent \textbf{Proof:} 
We observe that $\prod_{t=1}^T \mathcal N (x_t|O_1\ind_t, \Sigma) = \prod_{t=1}^T \mathcal N(x_t |O_2 \ind_t, \Sigma)$, if $( O_1 - O_2 )\ind_t =  0$, $\forall t \in [T]$, which is equivalent to $( O_1 - O_2 )R = \mathbf 0$. Due to Corollary \ref{cor:rankR}, $\textbf{dim}(\textbf{null}( R^\top)) \geq K -1$. Therefore we conclude that $( O_1 - O_2 )R = 0$ for $O_1 \neq O_2$.   $\hfill \square$

We also intuitively see the model is unidentifiable since there are $\nhmm \hmmK$ vectors to estimate in $O$ but we only have $\nhmm \hmmK - (\nhmm -1)$ linearly independent equations, as Corollary \ref{cor:rankR} suggests. Making this observation, we reduce the number of model parameters to $\nhmm \hmmK - (\nhmm -1)$ by setting a shared component $\dictel^k_{\hmmK} = s, \; \forall k \in [K]$, where $s \in \mathbb R^{\hmmL}$. 

\begin{defn}
\textbf{(The Shared Component Factorial Model - SC-FM)} The emission matrix of a SC-FM is of the form $\widetilde O = [\widetilde O^1, \dots,\widetilde O^k, \dots, \widetilde O^\nhmm, s ] $, where $\widetilde O^k \in \mathbb R^{L\times (\hmmK-1)}$, and $s \in \mathbb R^L$ is the shared component. The latent state indicators are either an indicator vector or an all zeros vector: $\widetilde r^k_t \in ( 0_{\hmmK-1} \cup e_{1:\hmmK-1}) $. The columns of the assignment matrix $\widetilde R$ are of the form $\widetilde R_t = [ (\widetilde r^1_t)^\top, \dots,(\widetilde r^k_t)^\top, \dots, (\widetilde r^\nhmm_t)^\top, K - \sum_{k=1}^K \sum_{m=1}^{\hmmK-1} \mathbf 1(\widetilde r^k_t = e_{m})  ]^\top$.  
\end{defn}

\begin{lem}
\label{lem:sciden}
{Let $\widetilde R^c \in \mathbb R^{(\nhmm \hmmK - (\nhmm -1)) \times \hmmK^\nhmm}$ denote a matrix whose columns consist of all possible combinations $\widetilde \ind_t$ can take (e.g. for $\hmmK=3$, $\nhmm =2$ case $R^c = \begin{bmatrix} e_1& e_1 &e_2 & e_2 & e_1 & e_2 & 0_2 & 0_2 & 0_2 \\ e_1 & e_2 & e_1 & e_2 & 0_2 & 0_2 & e_1 & e_2 & 0_2 \\ 0 & 0 & 0 &0 & 1 & 1 & 1 &1 & 2 \end{bmatrix}$ ). We conclude that $\textbf{rank}(\widetilde R^c) = \nhmm \hmmK - (\nhmm -1) $, and consequently $\textbf{rank}(\widetilde R) \leq \nhmm \hmmK - (\nhmm -1)$.} 
\end{lem}

 \noindent \textbf{Proof:} We will prove this by showing that the left null space of $\widetilde R$ only contains an all-zeroes vector. Let, 
$$\widetilde \indf^k_m(m_1, m_2,\dots,m_k,\dots,m_\nhmm) :=\left\{
     \begin{array}{lr}
       1, &  \text{if} \; m_k = m \\
       0,  &  \text{otherwise}  
     \end{array},
   \right. $$ and $q(m_1, m_2,\dots,m_k,\dots,m_\nhmm) := K - \sum_{k=1}^K \mathbf 1( m_k \neq 0 )$ for $k\in[K]$, $m \in [M-1]$ and $m_k \in 0 \cup [M-1]$. The first function represents the first $(\hmmK-1)\nhmm$ rows, and the second function represents the last row of $\widetilde R^c$, for the column that corresponds to the tuple $(m_1, m_2,\dots,m_k,\dots,m_\nhmm)$. 
For a vector $\alpha \in \mathbb R^{\nhmm \hmmK - (\nhmm -1)}$ in the left null space of $R^c$, $\alpha^\top R^c= 0_{\hmmK^\nhmm}$. Let us evaluate $\sum_{k,m} \alpha^k_m \widetilde \indf^k_m(m_1, m_2,\dots,m_k,\dots,m_\nhmm) +\alpha_q q(m_1, m_2,\dots,m_k,\dots,m_\nhmm)$ for the tuple $(0,\dots,0)$: 
\begin{align}
\sum_{k=1}^{\nhmm} \sum_{m=1}^{\hmmK-1} \alpha^k_m \widetilde \indf^k_m(0,\dots,0) + \alpha_{q} q(0,\dots,0) = K \alpha_{q}= 0 \label{eq:r1}
\end{align} 
So we conclude that $\alpha_{q}=0$. Next, we do the evaluation for the tuple $(0, \dots, m_k, \dots,0)$, where only one element $m_k$ is not equal to zero:
\begin{align}
\sum_{k=1}^{\nhmm} \sum_{m=1}^{\hmmK-1} \alpha^k_m \widetilde \indf^k_m(0, \dots, m_k, \dots,0)+ \alpha_{q} q(0, \dots, m_k, \dots,0) = \alpha^k_{m_k} +(K-1) \alpha_{q}. \label{eq:r2}
\end{align} 
By comparing Equations \eqref{eq:r1} and \eqref{eq:r2}, we see that $\alpha^k_{m} = 0$, $\forall k \in [K]$ and $\forall m \in [\hmmK-1]$. So, we conclude that $\textbf{dim}(\textbf{null}( (\widetilde R^c)^\top )) = 0$, and therefore from the rank-nullity theorem, $\textbf{rank}(\widetilde R^c) = \nhmm \hmmK - (\nhmm -1)$. And, if $\widetilde R$ contains all columns of $\widetilde R^c$ it has the same rank, which is the upper limit. $\hfill \square$

\begin{thm}
 {Given an assignment matrix $\widetilde R$ which contains all columns of $R^c$, the emission matrix of an SC-FM is identifiable. }
 \end{thm}
 
 \noindent \textbf{Proof:} After going through the same reasoning in Lemma \ref{lem:rank}, we again end up with the condition of having the term $( \widetilde O_1 - \widetilde O_2 )\widetilde R$ not equal to zero for two different emission matrices $\widetilde O_1 \neq \widetilde O_2$ for identifiability. As we have seen in Lemma \ref{lem:sciden}, $\textbf{dim}(\textbf{null}(\widetilde R^\top)) = 0$ in the case where $\widetilde R$ contains all possible assignment vectors. Therefore we conclude that $(\widetilde O_1 - \widetilde O_2 )\widetilde R \neq \mathbf 0$ for $\widetilde O_1 \neq \widetilde O_2$, and consequently the emission matrix of an SC-FM is identifiable, given an assignment matrix $\widetilde R$. $\hfill \square$

 
 
This theorem shows that the mapping $\widetilde O \to \widetilde X=\widetilde O \widetilde R$ is one-to-one. Even though this is the case, it is still not trivial to extract the columns of the emission matrix $\widetilde O$ from the observed data $\widetilde X$, simply because we do not have $\widetilde R$. However, we know the structure of $\widetilde R^c$, which contains all possibilities for the columns of $\widetilde R^c$. In the next section we will describe an algorithm which uses this fact.



\section{Learning}
\label{sec:learning}

What we propose for learning is the following: We first calculate an estimate for $\widetilde X^c$ with a clustering stage. Naturally, columns of $\widetilde X^c$ contains an arbitrary and an unknown permutation, which leads us to the system $\widetilde X^c \Pi = \widetilde O  \widetilde R^c \Pi$, where $\Pi \in \mathbb R^{\nhmm\hmmK \times \nhmm\hmmK}$ is a permutation matrix. This system has a different solution for different $\Pi$ matrices, and therefore we cannot solve this system for the true emission matrix unless we know $\Pi$. However, by assuming that the shared component $s$ is less correlated to the non-shared components than the correlation between the non-shared components, we will show that it is possible to extract the components by computing pairwise correlations between the columns of $\widetilde X^c$. 


To reduce the notation clutter we drop tilde's, although we still refer to the SC-FM parameters, and we use the regular factorial model notation where the indicator variable $r_t^k \in [M]$, for $k \in [K]$. Conforming with that notation we set the last columns of all the emission matrices to be the shared component, such that $\mu^k_\hmmK =s$, $\forall k \in [K]$. E.g., for $M=2$, $K=2$ case $O = [\mu_1^1, s, \mu_1^2,s]$.

\subsection{Learning the emission matrix from $X^c$}
In this section we describe an algorithm which extracts the columns of the emission matrix by looking at the pairwise correlations of the columns of $X^c$ matrix. The first step is to find which column of $X^c$ corresponds to the shared component.

\newcommand{\meanns}{x}

\newcommand{\mean}{x_l}
\newcommand{\meanp}{x_{l'}}
\newcommand{\meanpp}{x_{l''}}
\newcommand{\meanpone}{x^1}
\newcommand{\act}{r}

\begin{defn}
Let $\mean$ denote $l$'th column of $X^c$, so $\mean := X^c(:,l)=\sum_{k=1}^K \sum_{m=1}^{M-1} \mu^{k}_m \act^{k}_{m,l}+  \sum_{k=1}^K s \; \act^{k}_{M,l}$, where $\act^{k}_{m,l}, l \in [\hmmK^\nhmm]$ denotes the $m$'th entry of an indicator vector of length $\hmmK$ where only the $m$'th entry is one and the rest is zero, for the $k$'th emission matrix and $l$'th possible combination. 
\end{defn}


\begin{defn}
Let $v(\meanp) : \mathbb R^L \to \mathbb R^{M^K}$ denote a vector valued function with the argument $\meanp$, such that  
$v(\meanp) = \omega \left ( \left [  \left\langle \meanns_1,\meanp \right \rangle,  \left\langle \meanns_2,\meanp \right \rangle, \dots,  \left\langle \meanns_l,\meanp \right \rangle, \dots,  \left\langle \meanns_{\hmmK^\nhmm},\meanp \right \rangle \right ]  \right )$, where $\omega: [\hmmK^\nhmm  ] \to [\hmmK^\nhmm  ]$ is an ascending sorting mapping  such that $v_1(\meanp)\leq v_2(\meanp) \leq \dots \leq v_{M^K}(\meanp)$, where $v_l(\meanp)$ is the $l$'th smallest element in $v(\meanp)$ vector.  

\end{defn}



\begin{lem}
\label{lem:cor}
If $ \left \langle \mu^{k''}_{m''} ,s  \right \rangle  \leq  \left \langle \mu^k_m ,\mu^{k'}_{m'}   \right \rangle $, $\forall (k, k', k'') \in [K]$, and $\forall (m,m',m'')  \in [M-1]$, i.e. for any component $\mu^{k}_m$, the least correlated component is $s$, and $\left \langle \mu^{k}_{m},s \right \rangle \leq \left\langle s,s \right \rangle$, $\forall k \in [K]$, $m \in [M-1]$, i.e., the shared component $s$ has a non-trivial magnitude (e.g. all zeros vector doesn't satisfy this condition), then  
\begin{align}
Ks = \argmin\limits_{x_{l'},l' \in [\hmmK^\nhmm]} \sum_{l=1}^{(\hmmK-1)^\nhmm} v_l(x_{l'}),\; \text{for} \; M>2, K\geq 1.  
\end{align}

\end{lem}


\noindent \textbf{Proof Sketch:}
We want to show that given that the specified incoherence conditions are satisfied, the sum of the smallest $(\hmmK-1)^\nhmm$ terms in $\{\left \langle x_l, x_{l'} \right \rangle : l \in [\hmmK^\nhmm] \}$ get minimized when we set $x_{l'} = Ks$. In the proof given in supplemental material, we consider all possibilities for $x_{l'}$ and conclude that the minimizing possibility is $Ks$.

Lemma \ref{lem:cor} suggests that by computing pairwise correlations, it is possible to identify the column in $X^c$ which corresponds to $Ks$ component: The summation of first $(\hmmK-1)^\nhmm$ terms in $v(x_{l'})$  is minimized when we set $x_{l'} = Ks$. Therefore, we compute $v(x_{l'})$ for all columns of $X^c$, and assign the minimizing column to the term $Ks$. In $M=2$ case argmin of this summation contains multiple minimizers (including $Ks$), and we suggest a fix for that specific case with an additional assumption in the supplemental material. Now that we know how to estimate the $Ks$ term, next we look at the structure of $v(Ks)$ to extract the non-shared components. 

\begin{defn}
Let $\mathcal B_{K'}:= \{ l \in [\hmmK^\nhmm] : \sum_{k=1}^K \act^k_{M,l} =K' \} $, i.e. the indices $l$ for which $s$ appears $\nhmm-\nhmm'$ times, which corresponds to the terms of the form $ \sum_{k=1}^{\nhmm} \sum_{m=1}^{\hmmK-1} \mu^k_{m} r^k_{m,l} + (K-K')s$, $l \in \mathcal B_{K'}$. 
\end{defn}

\begin{lem}
\label{lem:nonshared}
Let $B_l^{K'} := \left \langle \sum_{k=1}^{\nhmm} \sum_{m=1}^{\hmmK-1} \mu^k_{m} r^k_{m,l} + (K-K')s, Ks  \right \rangle$, $l \in \mathcal B_{K'}$. If  $\left \langle s,\mu^k_m \right \rangle \leq \left\langle s,s \right \rangle $, $\forall k \in [K]$, and $\forall m\in [M -1]$, then for $\hmmK^\nhmm - (\hmmK-1)\nhmm \leq l' \leq \hmmK^\nhmm -1$ , $v_{l'}(Ks) = B_l^1$  for some $l\in \mathcal B_1$.

\end{lem}
\label{lem:bk}
\noindent \textbf{Proof:} Let us expand the expression $B_l^{K'}$: 
\begin{align*}
B_l^{K'}  = K\sum_{k=1}^{\nhmm} \sum_{m=1}^{\hmmK-1} \left \langle \mu^k_{m}, s  \right \rangle r^k_{m,l} + (K-K')K \left \langle s, s  \right \rangle, l \in \mathcal B_{K'}. 
\end{align*}
Since only $K'$ terms are active on the first term, and due to the condition $\left\langle s,s \right \rangle \geq \left \langle s,\mu^k_m \right \rangle $, $\forall k \in [K]$, $\forall m\in [M -1]$, we see that the above expression reaches the maximum value when $K' =0$. By the same token, we conclude that $B_l^{1} > B_{l'}^{K'}$, $\forall K' > 1$, $l\in \mathcal B_1$, $l' \in \mathcal B_{K'}$, since the number of $\left\langle s,s \right \rangle$ terms decrease as $ K' $ increases. Therefore, the largest elements of $v(Ks)$ after $v_{\hmmK^\nhmm}(Ks)$ correspond to $B_l^{1}$, $l\in \mathcal B_1$, as suggested by the lemma. $\hfill \square$   

We had an estimate for $s$ in the previous step, and now that we know which observed $x_l$ vectors correspond to the vectors comprised partly of $(K-1)s$  (i.e. terms corresponding to $\mathcal B_1$) from Lemma \ref{lem:bk}, we can estimate the non-shared components simply by subtracting $(K-1)s$ from each term in $\mathcal B_1$. The only remaining problem is to group them into proper emission matrices $O^{1:K}$. 

\subsubsection{Finding the grouping of the components}
We know from Lemma \ref{lem:cor} that the $(\hmmK-1)^\nhmm$ smallest elements of $v(Ks)$ (which also correspond to $\mathcal B_\nhmm$) are associated with all possible combinations of non-shared components that do not contain any term involving $s$. To find the groupings for the dictionary elements we solve a linear system of the form $Y= W H$ for $H$, where the columns of the $W$ matrix are the non-shared components estimated by subtracting $(K-1)s$ from components corresponding to $\mathcal B_1$, and columns of $Y$ correspond to all possible combinations of the non-shared components which correspond to $\mathcal B_\nhmm$. Solving this system figures out which combinations of the non-shared components corresponding to $\mathcal B_1$ add up to the combinations corresponding to $\mathcal B_\nhmm$, which are encoded in $H$. In practice we have observed that solving the following optimization problem which enforces sparsity on the columns of $H$ works well: $\widehat H = \argmin_H \| \widehat Y - \widehat W H \|_{F} + \sum_t\| H(:,t) \|_1 $. 


\subsubsection{Summary of emission matrix learning}
For a shared component factorial model (HMM or Mixture model), given the matrix of all possible observations $X^c \in \mathbb R^{L \times \nhmm^\hmmK}$, and provided that the columns of the emission matrix satisfy $\left \langle \mu^{k''}_{m''} ,s  \right \rangle  \leq  \left \langle \mu^k_m ,\mu^{k'}_{m'}   \right \rangle $, and $\left \langle \mu^{k''}_{m''},s \right \rangle \leq \left\langle s,s \right \rangle $, $\forall (k, k', k'') \in [K]$, $k\neq k'$ and $\forall (m,m',m'')  \in [M-1]$, Algorithm \ref{algo:estems} finds the columns of the emission matrix $O$ upto permutation among the columns of each emission matrix $O^k$ and permutation of the emission matrices.    

\begin{algorithm}
\caption{Emission matrix learning for F-GMM/F-HMM}
\label{algo:estems}
\begin{algorithmic}
\STATE  \textbf{Input:} The clustered data matrix $X^c \in \mathbb R^{L \times \nhmm^\hmmK}$
\STATE \textbf{Output:} Estimated emission matrix $\widehat O \in \mathbb R^{L \times \nhmm \hmmK}$  \\ 
\STATE $\bullet$ Compute the correlation matrix $C_{i,j} = \left \langle X^c(:,i), X^c(:,j) \right  \rangle$, $\forall i,j \in \mathbb R^{ \hmmK^\nhmm}$.
\STATE $\bullet$ Let $C^s$ denote the $C$ matrix with sorted rows in increasing order. Set $i^* = \argmin_i \sum_{j  = 1}^{(\hmmK-1)^\nhmm } C^s_{i,j}$, $v = C^s(:,i^*)$, and $\widehat s = X^c(:,i^*) / K$. 
\STATE $\bullet$ Find the indices of $(\hmmK-1)\nhmm$ largest elements in $v$, write the indices in $\mathcal B_1$. Set $\widehat W = X^c(:,\mathcal B_1) - (\nhmm-1)s 1^\top_{\nhmm-1}$.    
\STATE $\bullet$ Find the indices of $(\hmmK-1)^\nhmm$ smallest elements in $v$, write the indices in $\mathcal B_{\nhmm}$. Set $\widehat Y = X^c(:,\mathcal B_K) $.    
\STATE $\bullet$ Set $\widehat H = \argmin_H \| \widehat Y - \widehat W H \|_F + \sum_t\| H(:,t) \|_1 $, and group the columns of $\widehat W$ according to $\widehat H$ in $\widehat O$. 
\STATE $\bullet$ Output the corresponding estimate $\widehat O$.    
\end{algorithmic}
\end{algorithm}

\subsection{On Estimating $X^c$}
Even though the number of clusters $\hmmK^\nhmm$ is large, if the data is high dimensional then the initial clustering step can be done accurately. Let $d_{i,j} : =(X^c(:,i) + \epsilon_i) - (X^c(:,j)+\epsilon_j)$, where $\epsilon_i, \epsilon_j \sim \mathcal N(0,\sigma^2 I_{L})$. Notice $d_{i,j}$ is normally distributed such that, 
$$d_{i,j} \sim \mathcal N(X^c(:,i)-X^c(:,j), 2 \sigma^2 I ),$$ 
since $\epsilon_i, \epsilon_j$ are independent and spherical. Due to the concentration property of the Gaussians $\cite{dasgupta00}$ the distribution of $\|d_{i,j} - \mathbb E[d_{i,j}] \|_2^2$, will get concentrated around a thin shell of radius $\sqrt{2 L} \sigma$ such that, 
\begin{align}
\text{Pr}\left ( \left | \| d_{i,j} - \mathbb E[d_{i,j}] \|_2^2 - 2\sigma^2 L \right | >  c 2 \sigma^2 L  \right ) \leq 2 \exp(-Lc^2 /24),
\end{align}
where $c>0$ is a constant. This bound means that the magnitude of the noise on the pairwise distances between the true combinations $X^c$ gets bounded by $2\sigma^2 L$ for high dimensional data. Note that, in the case of correlated Gaussians the concentration property still holds around an elliptical shell \cite{dasgupta00}. A naive clustering approach such as running a randomly initialized k-means clustering can still fail, but a carefully crafted clustering algorithm such as \cite{dasgupta99} will return the true $X^c$ with high probability given that $\min_{i,j} d_{i,j} > \sigma \sqrt{L}$, and the smallest mixing weight is $\Omega(\frac{1}{\hmmK^\nhmm})$.  

\subsection{Estimating the auxiliary parameters}

\noindent \textbf{Hidden state parameters:} \\
Once we have an estimate $\widehat O$ for the emission matrix, the assignment matrix can be estimated by solving the optimization problem, $\widehat R = \argmin_{R} \| \widehat O R - X \|_F + \sum_{t=1}^T \| R(:,t) \|_1$. We estimate the assignment probabilities $\pi^{1:K}$ for F-GMM, or the transition matrices $A^{1:K}$ for F-HMM simply by counting the occurrences in $\widehat R$: \\
$\widehat \pi_{i}^k = \frac{1}{T} \sum_{t=1}^T \mathbf 1(\widehat r^k_t = e_i ),\;  \widehat A^k_{i,j} = \frac{1}{T-1}\sum_{t=1}^{T-1} \mathbf 1(\widehat r^k_{t+1} = e_i )  \mathbf 1(\widehat r^k_{t} = e_j ),\; i,j \in  [M], k \in [K]
$.
In practice, $\widehat R$ is noisy and the entries are not binary. We threshold the $\widehat R$ matrix to make it binary before the counting step.  \\ 

\noindent \textbf{Covariance matrix:} \\
Once we have estimates for the emission and the assignment matrix, we subtract the reconstruction from the data to make it zero mean. After that the covariance matrix is estimated with the usual covariance estimator: $\widehat \Sigma = \frac{1}{T-1}\sum_{t=1}^T \left (X_t - (\widehat O \widehat R)_t  \right ) \left (X_t - (\widehat O \widehat R)_t \right )^\top$, where $(\widehat O \widehat R)_t$ denotes the reconstruction at time $t$.



\section{Experiments}

\subsection{Synthetic Data}
We conducted experiments with synthetic data generated from shared component factorial model. We set $M=3$ and $K=2$. The columns of the emission matrix are sampled from a Gaussian with variance 10. The observation noise variance $\sigma^2$, data dimensionality $L$, and number of observations $T$ were all varied to compare the behavior of the proposed approach and EM. For the clustering step in the proposed approach, we applied the algorithm in \cite{dasgupta99}. For EM, we used 10 restarts with dictionaries started at the perturbed versions of the mean of the observed data. We report the result of the initialization that resulted in the highest likelihood. As error, we report the euclidean distance between the estimated dictionary matrix $O$ and the true dictionary, by resolving the permutation ambiguity. Figure \ref{fig:synth_exps} shows various comparisons between the two algorithms in terms of accuracy in recovering the true dictionaries and run time. The parameter setup for the fixed variables is shown under each figure. We see that the algorithm works much better than EM in general. We also see from Figure \ref{fig:synth4} that the proposed approach is faster, and potentially more scalable than EM. 

\newcommand{\sz}{.245}

\begin{figure}[t]
\centering
\begin{subfigure}[b]{\sz\textwidth}
                \includegraphics[width=\textwidth]{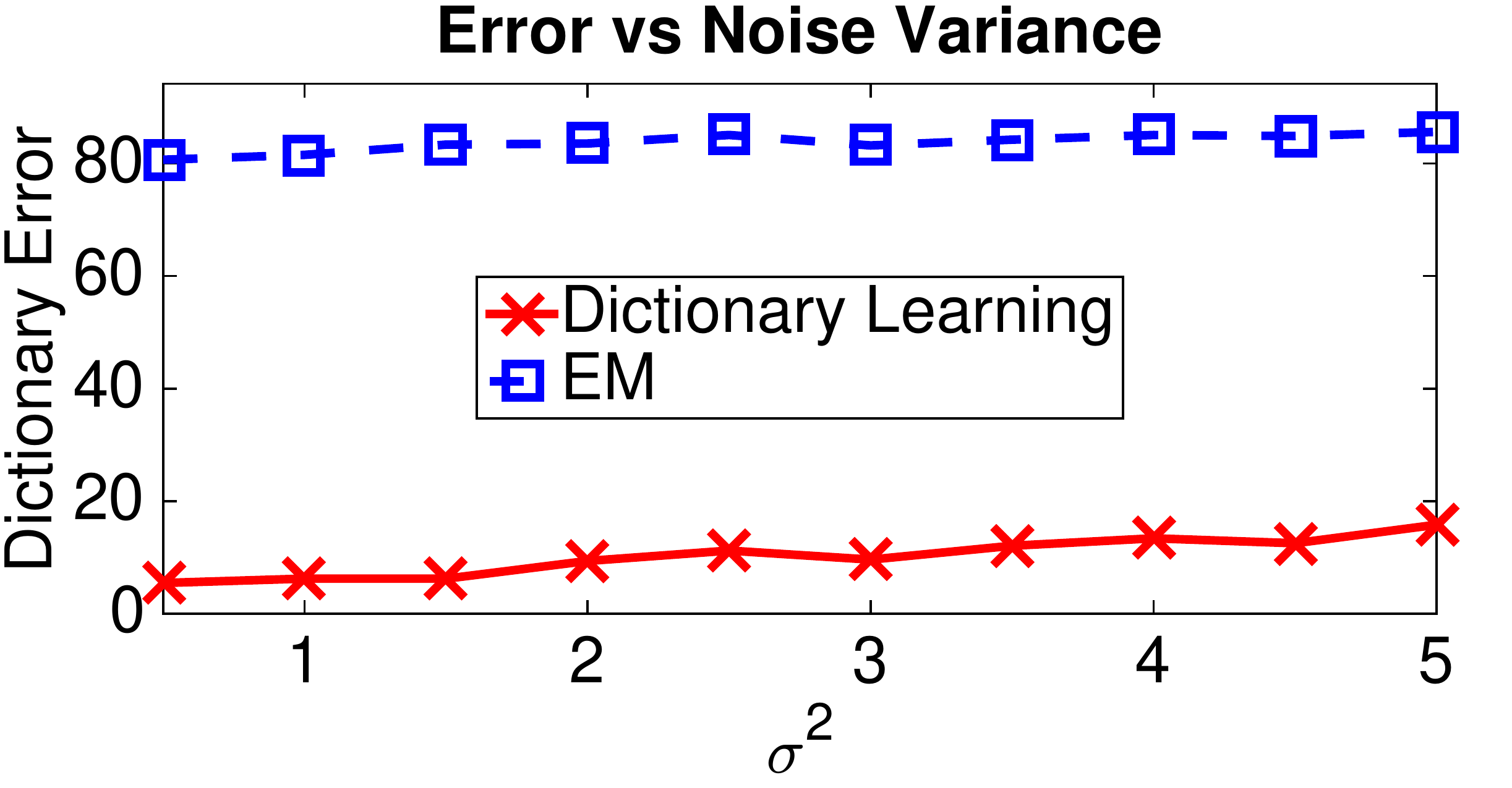}
                \caption{Error vs $\sigma^2$, $L = 50$, $T= 200$.}
                \label{fig:synth1}
\end{subfigure}
\begin{subfigure}[b]{\sz\textwidth}
                \includegraphics[width=\textwidth]{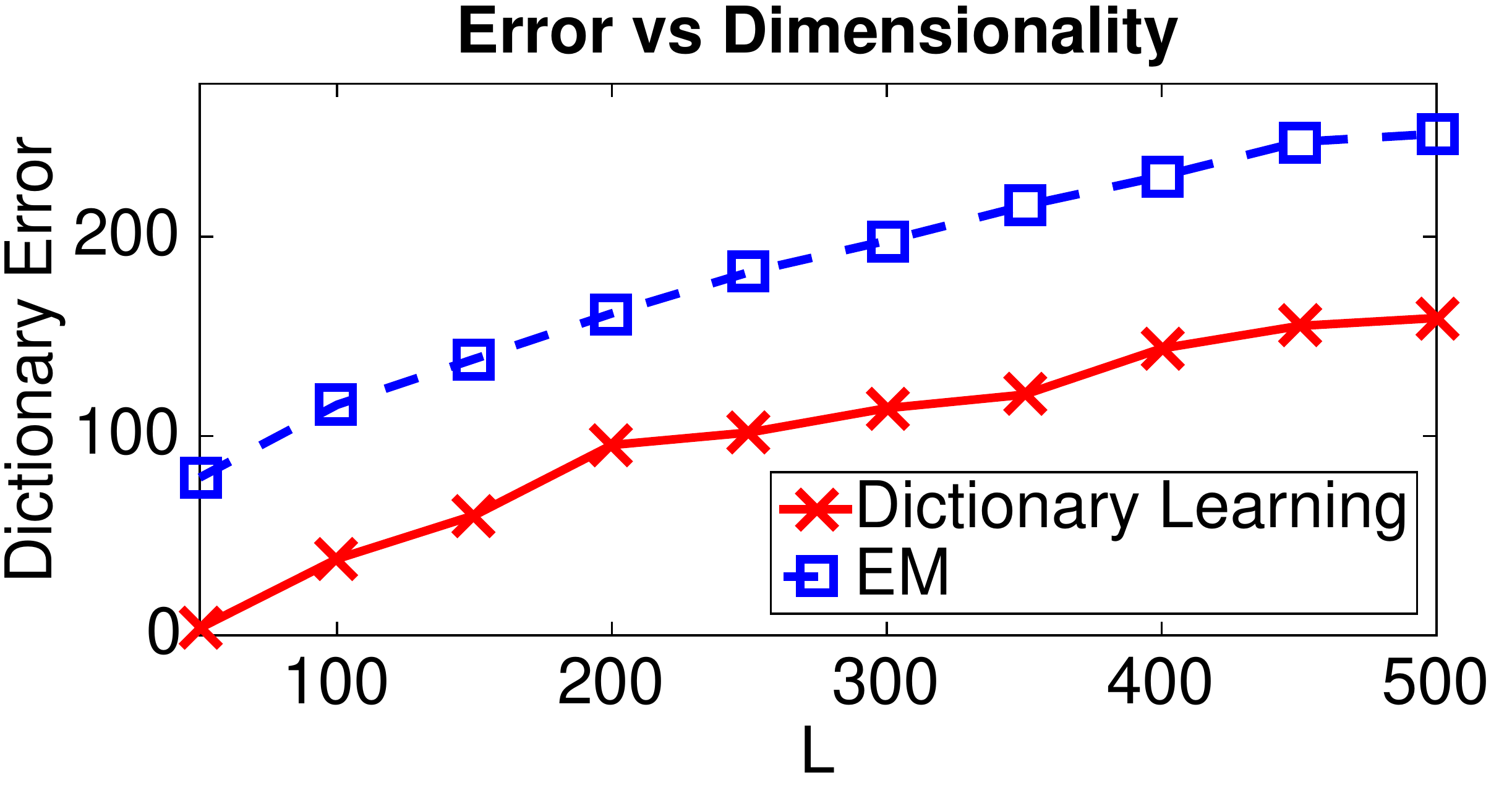}
                \caption{Error vs $L$, $\sigma^2 =0.5$, $T=200$.}
                \label{fig:synth2}
\end{subfigure}
\begin{subfigure}[b]{\sz\textwidth}
                \includegraphics[width=\textwidth]{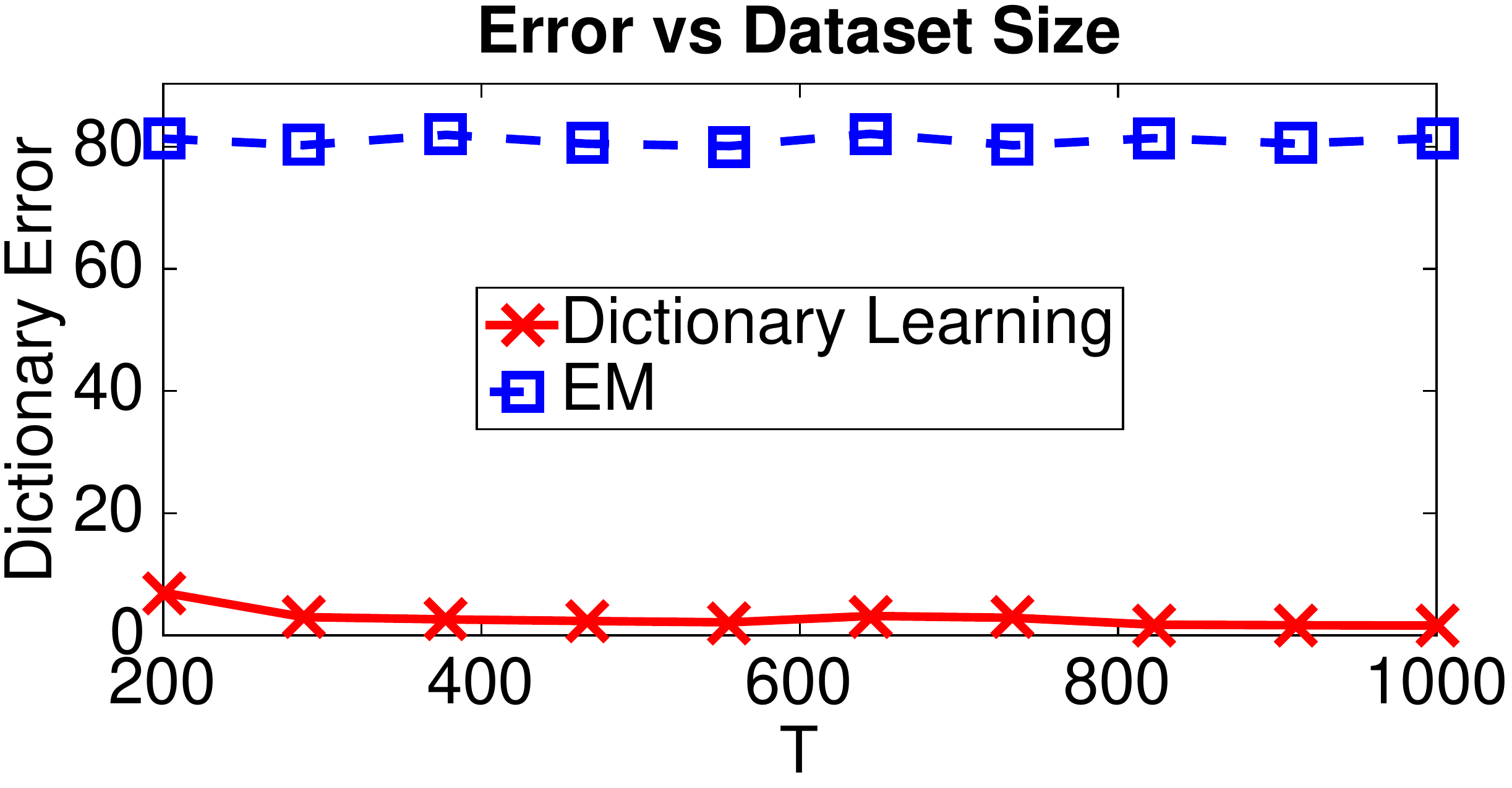}
                \caption{Error vs $T$, $L=50$, $\sigma^2 =0.5$. }
                \label{fig:synth3}
\end{subfigure}
\begin{subfigure}[b]{\sz\textwidth}
                \includegraphics[width=\textwidth]{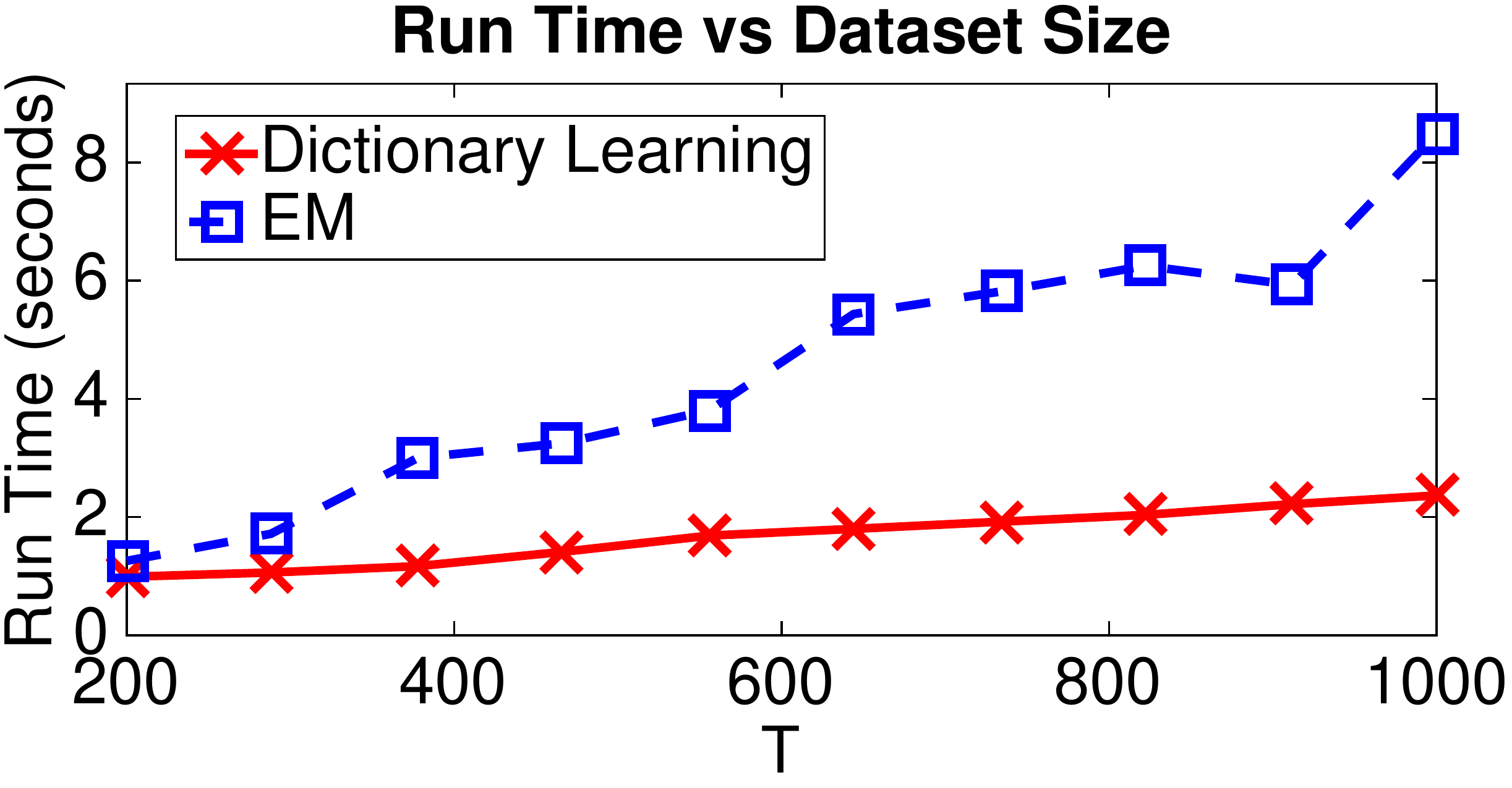}
                \caption{Run time vs $T$, $L=50$, $\sigma^2 =0.5$.}
                \label{fig:synth4}
\end{subfigure}
\caption{Various performance measures for the proposed algorithm and EM on synthetic data averaged over $50$ trials.}
\label{fig:synth_exps}
\end{figure}

\subsection{Digit Data}
\begin{figure}[t]
\centering
\begin{subfigure}[b]{0.7\textwidth}
                \includegraphics[width=\textwidth]{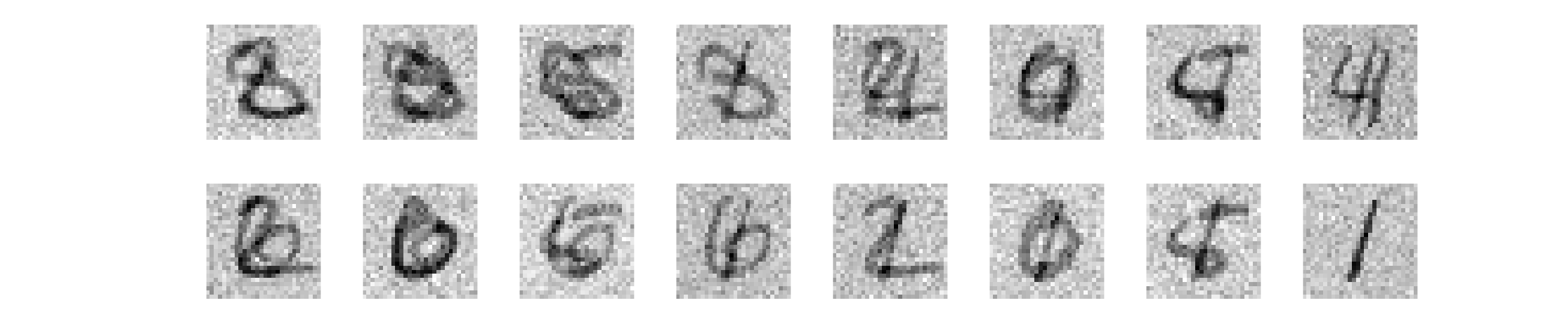}
                \caption{All possible combinations for the observations}
                \label{fig:combs}
\end{subfigure}%

\begin{subfigure}[b]{.45\textwidth}
                \includegraphics[width=\textwidth]{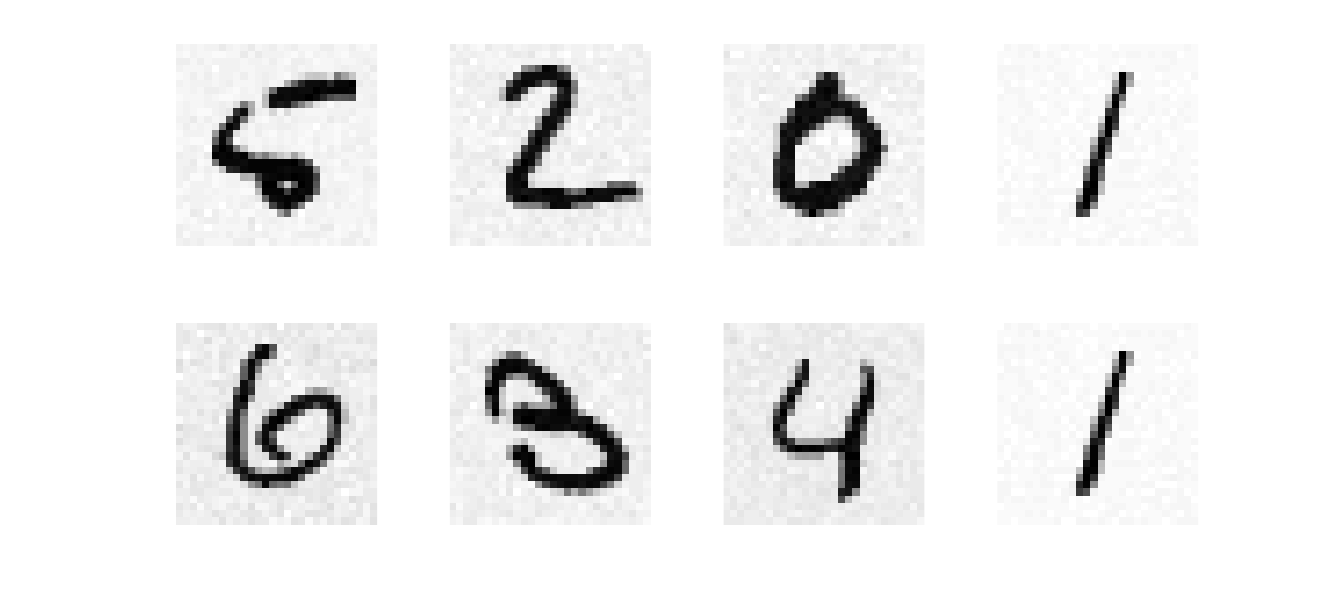}
                \caption{Dictionary Learning}
                \label{fig:factors}
\end{subfigure}
\begin{subfigure}[b]{.45\textwidth}
                \includegraphics[width=\textwidth]{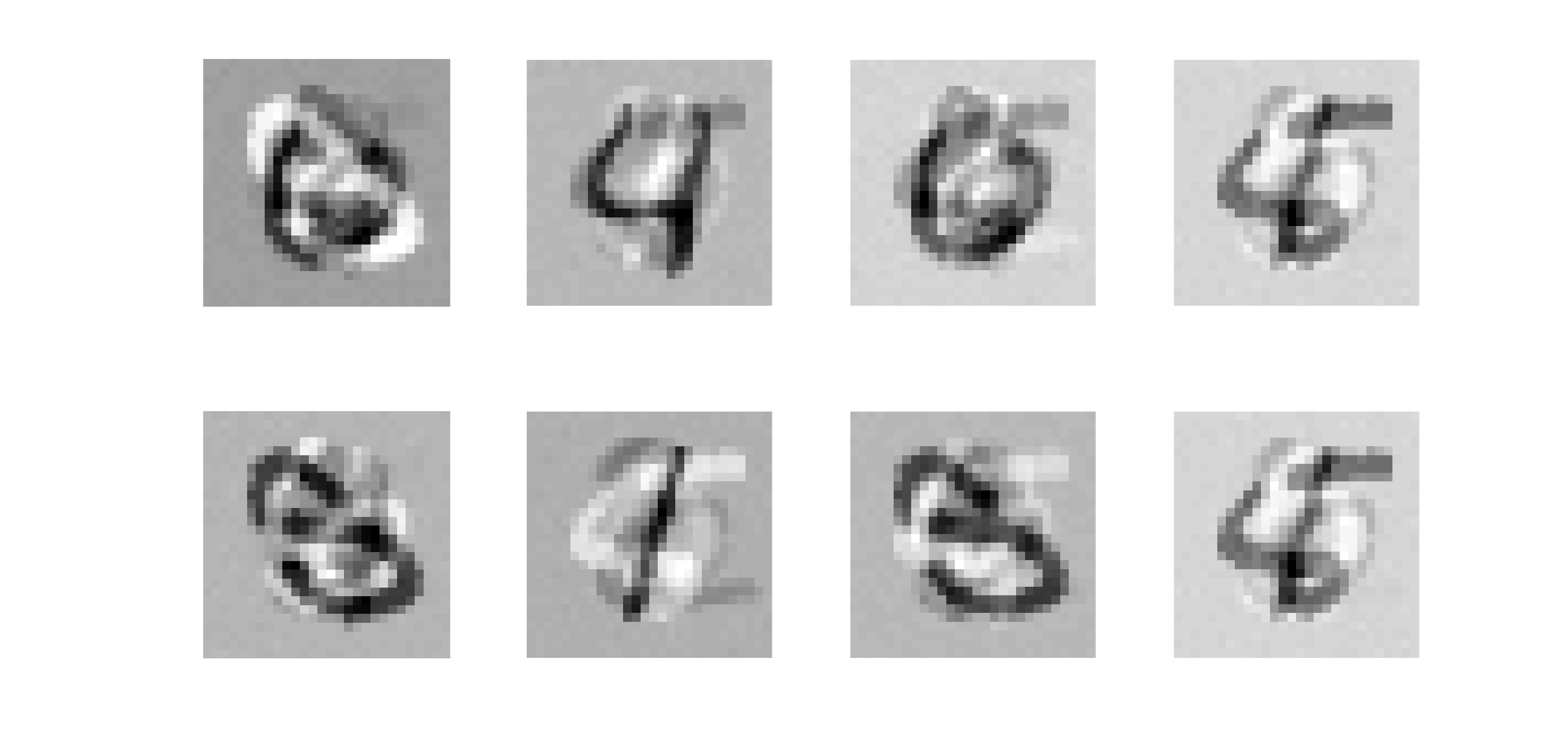}
                \caption{EM}
                \label{fig:factors}
\end{subfigure}

\caption{Unmixing of synthetically mixed noisy digit images with SC-FM. Figures (b) and (c) show the the learned emission matrices for the proposed algorithm and EM. A row in Figures (a) and (b) corresponds to the components corresponding to the same group. }
\label{fig:digits}
\end{figure}

In this experiment, we work with digit images from the MNIST dataset. We compare the proposed dictionary learning approach in Section \ref{sec:learning} with an EM algorithm, on synthetically combined images according to the shared component factorial model, where we set $\hmmK = 4$, and $\nhmm = 2$. We generate 2000 such images. The images are of size $28 \times 28$. We normalize the pixel values so that they take on values between $0$ and $1$. We add spherical Gaussian noise with standard deviation $\sigma = 0.22$ to every generated image. We initialize the columns of the emission matrix in EM with the randomly perturbed versions of the mean of the generated data. We do 10 such random initializations and pick the initialization with the highest likelihood. In Figure \ref{fig:digits}, we show the all noisy versions of $16$ possible combinations. We also show the reshaped versions of the learned columns of the dictionaries for the proposed algorithm and EM. 

We see that the estimates obtained with dictionary learning approach are close to the true digits, whereas EM finds a local solution which deviates from the true digits significantly.

\section{Conclusions and Discussion}
In this paper we have shown that the standard factorial model in the literature is not learnable. We then proposed an exact algorithm for the case where there is a one column sharing assumption between $\nhmm$ emission matrices. Although we have focused on the one component sharing case in this paper, it is possible to derive algorithms for multiple component sharing cases, under certain incoherence assumptions as future work. One other interesting future direction is to derive a learning procedure which would be able to extract the model parameters with fewer outputs from the clustering stage: We have shown in Section \ref{sec:identify} that the number of linearly independent combinations in $R$ is $\hmmK\nhmm -(\nhmm -1)$, which is much smaller than the number of all possible combinations $\hmmK^\nhmm$. The challenge would be to identify the correspondences between the observed vectors and the actual combination they are associated with.



\bibliographystyle{../../misc/bibfiles/fbe_tez_v11.bst}
\bibliography{../../misc/bibfiles/ref.bib}

\end{document}